\documentclass[10pt,twocolumn,letterpaper]{article}

\usepackage{wacv}              

\usepackage{graphicx}
\usepackage{amsmath}
\usepackage{amssymb}
\usepackage{booktabs}
\usepackage{enumitem}
\setlist[itemize]{leftmargin=3mm}

\usepackage[pagebackref,breaklinks,colorlinks]{hyperref}

\usepackage[capitalize]{cleveref}
\crefname{section}{Sec.}{Secs.}
\Crefname{section}{Section}{Sections}
\Crefname{table}{Table}{Tables}
\crefname{table}{Tab.}{Tabs.}

\def\wacvPaperID{2703} 
\def\confName{WACV}
\def\confYear{2025}

\begin{document}

\title{Behaviour4All: in-the-wild Facial Behaviour Analysis Toolkit}

\author{Dimitrios Kollias\\
Queen Mary University of London\\
{\tt\small d.kollias@qmul.ac.uk}
\and
Chunchang Shao\\
Queen Mary University of London\\
\and
Odysseus Kaloidas\\
London School of Economics\\
\and
Ioannis Patras\\
Queen Mary University of London\\
}
\maketitle

\begin{abstract}
In this paper, we introduce Behavior4All, a comprehensive, open-source toolkit for in-the-wild facial behavior analysis, integrating Face Localization, Valence-Arousal Estimation, Basic Expression Recognition and Action Unit Detection, all within a single framework.
Available in both CPU-only and GPU-accelerated versions, Behavior4All leverages 12 large-scale, in-the-wild datasets consisting of over 5 million images from diverse demographic groups. It introduces a novel framework that leverages distribution matching and label co-annotation to address tasks with non-overlapping annotations, encoding prior knowledge of their relatedness.
In the largest study of its kind, Behavior4All outperforms both state-of-the-art and toolkits in overall performance as well as fairness across all databases and tasks. It also demonstrates superior generalizability on unseen databases and on compound expression recognition. Finally, Behavior4All is way times faster than other toolkits.
\end{abstract}

\vspace{-0.5cm}





\section{Introduction}


Understanding human behaviour can be approached in several ways. One common method is the detection of facial muscle movements, known as Action Units (AUs), which are systematically categorized by the Facial Action Coding System (FACS) \cite{ekman1997face}. This approach focuses on the granular analysis of facial muscle activity, providing a detailed understanding of how specific muscles contribute to different expressions.
Another approach involves interpreting the emotional message conveyed by a facial expression, where the expression is linked to a particular emotional state. This method is often employed in basic expression recognition, which classifies facial expressions into fundamental categories (happiness, sadness, etc).
Finally, valence-arousal estimation \cite{russell1980circumplex} is a crucial concept in emotion analysis. Valence refers to the positivity or negativity of an emotion, while arousal measures the intensity of the emotion. By estimating these two dimensions, we can better understand the subtleties of emotional expression beyond basic categories, capturing a more nuanced picture of human behaviour. Together, these methods (AU Detection, AUD; Basic Expression Recognition, BER; Valence-Arousal Estimation, VA-E) offer comprehensive tools for decoding the rich tapestry of emotions expressed through facial movements.


Over the past two decades, researchers have increasingly focused on Automatic Behaviour Analysis (ABA), a critical processing step in a wide range of applications, including ad testing, driver state monitoring, HCI and healthcare.
Several architectures have been developed for ABA, with deep learning (DL) methods demonstrating promising performance. In recent years, some toolkits for ABA have emerged. 
However, the datasets used to train these architectures and toolkits present several significant limitations.
Firstly, they are captured in controlled conditions (e.g., with limited illumination and fixed camera angle), hampers the robustness of the resulting models when applied to naturalistic, unconstrained (termed 'in-the-wild') conditions. Secondly, these datasets often feature a relatively small number of subjects, making the models susceptible to overfitting and limiting their generalizability. Thirdly, the demographic diversity of these datasets is typically narrow, leading to models that perform suboptimally on under-represented demographic groups. 

Furthermore, most datasets are annotated for only a single  task, which has led to the predominance of single-task models over multi-task (MT) ones. Consequently, existing toolkits depend on separate models for each behaviour task, with these models typically trained on a single database. Even in cases where MT models have been developed, the risk of negative transfer \cite{wang2019characterizing} may arise, potentially compromising their performance and generalizability.
Additionally, all existing toolkits are not performing valence-arousal estimation.
Lastly, some of the toolkits (e.g., OpenFace, OpenFace 2.0 \cite{baltruvsaitis2016openface}, and py-feat \cite{cheong2023py}) rely on traditional machine learning methods (e.g., SVM, HOG, XGB, PCA), which are less accurate compared to contemporary DL models. 

\begin{table*}[h]
\caption{Comparison of facial behaviour analysis tools} 
\label{toolkits}
\centering
\scalebox{.65}{
\begin{tabular}{ |c||c|c|c|c|c|c|c|c|c|c|c|c|c|c| }
 \hline
Toolkit & Face Bbox & Landmarks & AU & BER & VA & Real-time & Free & GPU support & Train & Test & MTL & in-the-wild dbs & multiple dbs & Downstream Tasks\\ 
  \hline
 \hline
AFFDEX 2.0 & \checkmark & \checkmark & \checkmark & \checkmark &  &
\checkmark &  & &  &  &  & \checkmark & & \\
\hline
FACET & \checkmark & \checkmark & \checkmark & \checkmark &  & \checkmark &  &  &  &  &  & &  &\\
\hline
\hline
OpenFace 2.0 & \checkmark & \checkmark & \checkmark &  &  & \checkmark & \checkmark &  & \checkmark & \checkmark &  & &\checkmark & \\
\hline
LibreFace  &    &   & \checkmark  & \checkmark  &  & \checkmark & \checkmark & \checkmark & \checkmark & \checkmark &  & \checkmark &  & \\
\hline
py-feat &  \checkmark  & \checkmark   & \checkmark  & \checkmark  &  & \checkmark & \checkmark & \checkmark &  & \checkmark &  & \checkmark & \checkmark &  \\
\hline
\hline
\begin{tabular}{@{}c@{}} \textbf{Behaviour4All}  \end{tabular}    &   \checkmark  & \checkmark   & \checkmark  & \checkmark  & \checkmark & \checkmark & \checkmark & \checkmark & \checkmark & \checkmark & \checkmark & \checkmark & \checkmark & \checkmark \\
\hline
\end{tabular}
}
\end{table*}

ABA is lacking an accurate, fair, efficient, open-source, real-time and standalone toolkit that is capable of performing the different ABA tasks (Face Detection, Face Alignment, AUD, BER and VA-E).
In this paper, we build a toolkit named Behaviour4All for in-the-wild Facial Behaviour Analysis.
Behaviour4All addresses the aforementioned challenges in ABA by offering a comprehensive solution capable of performing multiple ABA tasks while overcoming the limitations highlighted above. Behavior4All is composed of 2 primary components: FaceLocalizationNet and FacebehaviourNet. The first one performs simultaneous face detection and landmark localization; the second performs simultaneous 17 AU Detection,  7 Basic Expressions Recognition and VA Estimation. 

FaceLocalizationNet is a single-stage DL face detector that utilizes a feature pyramid network, producing five feature maps at different scales to detect both large and small faces. It also includes a context head module that processes a feature map at a specific scale and computes a cascaded multi-task loss, capturing more contextual information surrounding the faces.
FacebehaviourNet is a CNN designed for Multi-Task Learning (MTL), structured around residual units. During model training, co-training through task relatedness, derived from prior knowledge, and distribution matching are employed to effectively aggregate knowledge across datasets and transfer it across tasks. This approach is particularly beneficial when dealing with non-overlapping annotations, as it enhances the model's performance and mitigates the risk of negative transfer.
Our major contributions are summarized as follows:


\begin{itemize}[noitemsep,nolistsep]
    
\item We introduce Behavior4All, a comprehensive open-source toolkit designed for accurate and efficient real-time facial behavior analysis, available in CPU-only and GPU-accelerated versions. Behavior4All is the first toolkit to integrate the following functionalities (especially the 3 behaviour tasks): Face Detection and Alignment, Valence-Arousal Estimation, Basic Expression Recognition, and AU Detection, all performed simultaneously within a single framework.

\item  For training and testing our toolkit, we employ 12 large-scale in-the-wild datasets comprising over 5 million images, featuring participants from diverse demographic groups. We propose a novel framework that leverages distribution matching and label co-annotation for tasks with non-overlapping annotations, incorporating prior knowledge of their relatedness into the encoding process.

\item We conduct an extensive experimental study, the largest of its kind, to the best of our knowledge. In this study, at first, 
we compare both the overall performance and fairness of our toolkit across 8 databases against state-of-the-art (sota) and existing toolkits (OpenFace, LibreFace \cite{chang2024libreface} and py-feat). Our toolkit not only outperforms all sota and toolkits across all databases and tasks, but also exhibits greater fairness. Notably, our toolkit is often considered fair across various demographic groups. Next,
we evaluate the generalizability of our toolkit on 4 unseen databases and for compound expression recognition, where our toolkit surpasses all sota.
Finally, we assess the computational cost of our toolkit in comparison to other toolkits. Behavior4All runs at least 1.9 times faster than OpenFace and py-feat, and while achieving similar efficiency to LibreFace.

\end{itemize}

\section{Related Work}

\paragraph{Toolkits} In recent years, some toolkits for facial behaviour analysis have been developed. 
Table \ref{toolkits} presents an overview of these toolkits.
\emph{Face Bbox} indicates whether face detection is performed as part of the toolkit, 
or if any external face detection software is used. 
\emph{Landmarks} refer to whether landmark localization is performed. Landmarks are essential for face alignment. 
\emph{AU/BER/VA} indicate whether AUD/BER/VA-E is performed.
%
%
%
%
\emph{Free} indicates whether the tool is freely available for research purposes.
%
%
\emph{Train/Test} indicate the availability of model training source code and of checkpoints and codes for inference. 
\emph{MTL} indicates whether one Multi-Task model is provided for all tasks. Let us note that our toolkit is the only one that provides one model that simultaneously addresses all 3 behaviour tasks; all other toolkits have a separate model for each task, whilst tackling at max 2 tasks (rather than 3). 
\emph{in-the-wild dbs} indicate whether in-the-wild databases have been used in the development of the toolkit. Let us mention that our toolkit is the only one that utilizes an in-the-wild database in the AUD task, whilst using only in-the-wild databases for all other tasks.
\emph{multiple dbs} indicate whether multiple databases have been used in the development of the toolkit. 
\emph{Downstream Tasks} indicate whether the toolkit has shown its premise into downstream tasks or if its generalizabilaity has been tested in other datasets.

\begin{figure*}[h]
\centering
\includegraphics[height=4.7cm]{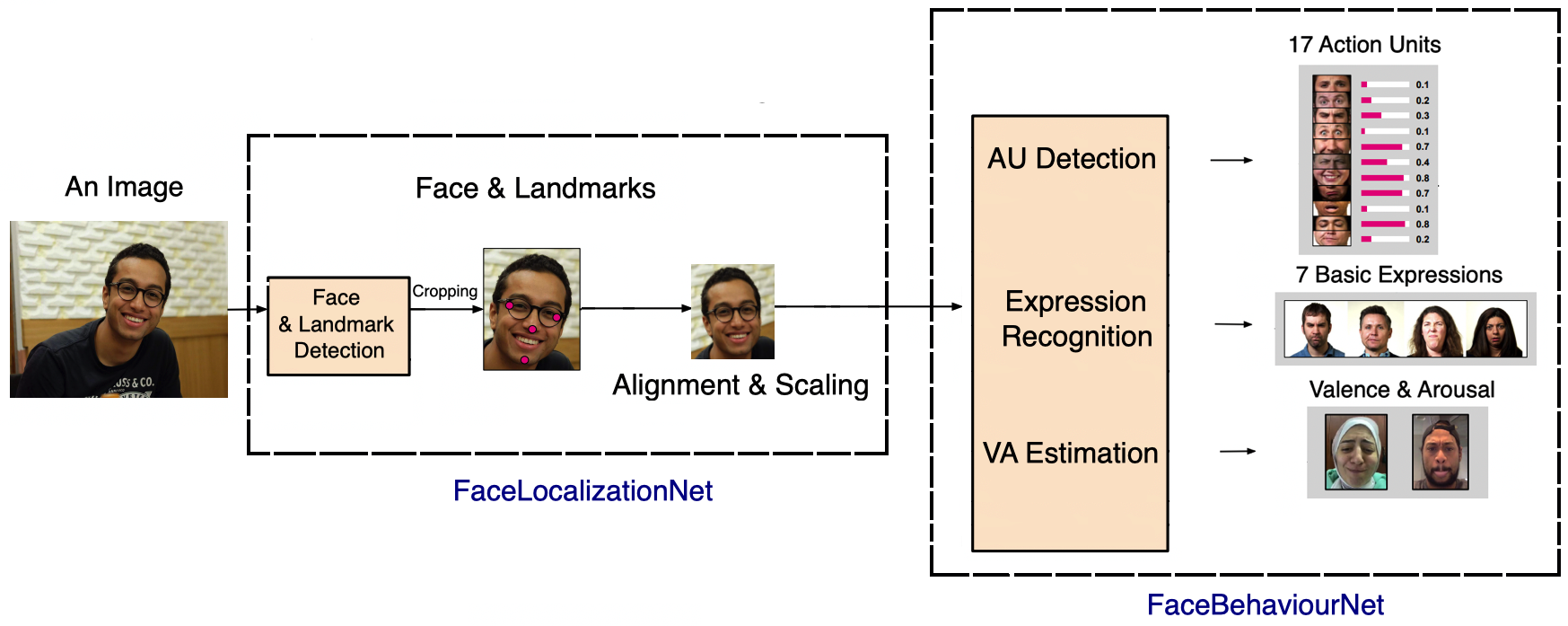} 
\caption{The full pipeline of Behaviour4All Toolkit}
\label{overview}
\end{figure*}

\noindent
\textbf{State-of-the-art} 
  DAN \cite{dan} consists of 3 components that enhance class separability, while focusing on multiple facial regions simultaneously. MA-Net \cite{manet} combines a multi-scale module to enhance feature diversity and robustness with a local attention module that focuses on salient facial regions. EAC \cite{eac} improves BER under noisy labels by using attention consistency combined with random erasing to prevent the model from memorizing noisy samples. 
  ME-GraphAU \cite{megraphau} employs a multi-dimensional edge feature-based AU relation graph that learns the relationships between pairs of AUs. AUNets \cite{aunets} predicts the viewpoint of a video first and then applies an ensemble of AU detectors specifically trained for that viewpoint. Res50, the winner of Emotionet Challenge, employed a ResNet50 and extra automatically annotated images.
  AffWildNet \cite{kollias2019deep} is an end-to-end CNN-RNN, which integrates facial landmarks within the network's design, trained with a correlation-based loss and employing a rebalancing data augmentation strategy.  
  VA-StarGan is a VGG16 trained with both real and generated images of various VA states.
  MT-EmotiEffNet \cite{emogcn} is a MTL framework leveraging EfficientNet as a backbone to jointly learn BE, AUs, and VA. 
FUXI \cite{fuxi}, SITU \cite{situ} and CTC \cite{ctc} are the top-3 best performing methods on the ABAW Competition for BER, AUD and VA-E, that extract multi-modal features and combine them with Transformers.

\section{Behaviour4All Toolkit}


Figure \ref{overview} shows an overview of Behaviour4All, which is composed of two primary components: FaceLocalizationNet and FacebehaviourNet. The first component performs image pre-processing, which involves simultaneous Face and Landmark Detection and Image Alignment. The pre-processed images are then fed to FacebehaviourNet, performs simultaneous Detection of 17 AUs, Recognition of 7 Basic Expressions and Estimation of Valence-Arousal. These components are explained in detail in the
following.




\vspace{-0.3cm}
\paragraph{FaceLocalizationNet}



For face detection and alignment we used RetinaFace \cite{deng2020retinaface}, which is a state-of-the-art deep learning-based face detector designed for robust and high-accuracy face detection. The key idea behind RetinaFace is to detect faces with a high level of precision, even under challenging conditions such as variations in lighting, pose, occlusion, and scale.
RetinaFace is a single-stage face detector, meaning that it predicts face locations and key points directly from the image without requiring a two-step process (e.g., region proposal and refinement). This makes it faster and more efficient compared to two-stage detectors like Faster R-CNN. 
RetinaFace consists of three main components:
the feature pyramid network, the context head module and the cascade multi-task loss. 

First, the feature pyramid network gets the input face images and outputs five feature maps of different scale so as to enable the detection of both large and small faces. It uses a backbone network to extract features from the input images, which is typically a ResNet. In our case, we employ both a MobileNet model, as well as a ResNet model.
Then, the context head module gets a feature
map of a particular scale and calculates a multi-task loss.  The context head module helps in capturing more contextual information around faces, improving the detection performance in cluttered environments.
RetinaFace simultaneously detects face bounding boxes, 5 (key) facial landmarks (two eyes, nose tip, and mouth corners), and provides a 3D position estimation. The multi-task approach helps in improving the accuracy of face detection by leveraging related tasks. The landmark localization branch (that predicts five facial landmarks)  is particularly useful for face alignment and behaviour analysis tasks, as we will explain in a bit.

In our toolkit, we provide four different versions of this module, which we have trained both on Tensorlow and on Pytorch libraries, using the Wider Face dataset which is a standard dataset containing many in-the-wild images with a high degree of variability in scale, pose, expression, occlusion and illumination. One version is RetinaFace with ResNet-50 as backbone network; another version is with MobileNet-0.25 as backbone; and two final versions that are quantized versions of these two cases; the quantized versions are lighter models that maintain quite similar performance to the originals. 

Prior to inputting a face detected image into the next module, it is imperative to perform facial image alignment based on the 5 localized facial landmarks of RetinaFace. Facial image alignment involves geometric transformations, such as translation, rotation, and scaling, to convert the input face image into a
canonical or standardized form. This process ensures consistent positioning of facial features across various images, facilitating the learning of patterns by our module.


Once both the facial crop and the face alignment have been performed, the aligned faces are scaled to a fixed size of $112 \times 112 \times 3$, and passed as an input to
the next module.

\subsection{FacebehaviourNet}

FacebehaviourNet is a Multi-Task Learning (MTL) CNN model that concurrently performs: (i) continuous affect estimation in terms of Valence and Arousal (VA); (ii) recognition of 7 basic facial expressions; and (iii) detection of activations of $17$ binary facial Action Units (AUs). 

For a given image, we can have label annotations of either one of seven basic expressions $y_{expr} \in \{1,2,\ldots,7\}$, or $17$\footnote{In fact, $17$ is an aggregate of action units in all datasets; typically each dataset has from 10 to 12 AUs} binary AU activations $y_{AU} \in \{0,1\}^{17}$, or two continuous affect dimensions, valence ($y_{V} \in [-1,1]$) and arousal ($y_{A} \in [-1,1]$).

%
We train FacebehaviourNet by minimizing the objective function: $\mathcal{L}_{MT} =$  
\begin{align}
 \lambda_{1} \mathcal{L}_{Expr} + \lambda_{2} \mathcal{L}_{AU} + \lambda_{3} \mathcal{L}_{VA}  + \lambda_{4} \mathcal{L}_{DM} + \lambda_{5} \mathcal{L}_{SCA} 
\label{eq:mt1}
\end{align}
\noindent where: $\mathcal{L}_{Exp}$ is the cross entropy (CE) loss computed over images with basic expression label; 
$\mathcal{L}_{AU}$ is the binary CE loss computed over images with AU activations;
$\mathcal{L}_{VA} = 1 - 0.5 \cdot ({CCC}_A + {CCC}_V)  $ is the Concordance Correlation Coefficient (CCC) based loss computed over images with VA labels;
$\mathcal{L}_{DM}$ and $\mathcal{L}_{SCA}$ are the distribution matching and soft co-annotation losses, which are derived based on the relatedness between expressions and AUs. The derivation of these losses is detailed in the subsequent sections. 

 The two losses are essential for model training due to the non-overlapping nature of the utilized databases' task-specific annotations. For instance, one database only includes AU annotations, lacking valence-arousal and 7 basic expression labels.
 Training the model directly with these databases using a combined loss function for all tasks would result in noisy gradients and poor convergence, as not all loss terms would be consistently contributing to the overall objective function. This can lead to issues typical of MTL, such as task imbalance (where one task may dominate training), or negative transfer (where the MTL model underperforms compared to single-task models) \cite{li2023identification}. Finally, these two losses aim to ensure consistency of the model's predictions between the different tasks.

\paragraph{\textbf{Task-Relatedness}} 
The study by \cite{du2014compound} conducted a cognitive-psychological analysis of the associations between facial expressions and AU activations, summarizing the findings in Table \ref{relationship} that details the relatedness between expressions and their corresponding AUs. Prototypical AUs are those consistently identified as activated by all annotators, while observational AUs are those marked as activated by only a subset of annotators.

\begin{table}[h]
\centering
\caption{
Relatedness of expressions \& AUs inferred from \cite{du2014compound}; in parenthesis are the weights that denote fraction of annotators that observed the AU activation
}
\scalebox{0.8}{
\begin{tabular}{|l|c|c|}
\hline
Expression   & Prototypical AUs & Observational AUs \\
\hline\hline
happiness &  12, 25 & 6 (0.51)   \\
\hline
sadness &  4, 15 & 1 (0.6),\hspace{0.08cm} 6 (0.5),\hspace{0.08cm} 11 (0.26),\hspace{0.08cm} 17 (0.67) \\
\hline
fear &  1, 4, 20, 25 &2 (0.57),\hspace{0.08cm} 5 (0.63),\hspace{0.08cm} 26 (0.33) \\
\hline
anger &4, 7, 24 &10 (0.26),\hspace{0.08cm} 17 (0.52),\hspace{0.08cm} 23 (0.29) \\
\hline
surprise &1, 2, 25, 26 &5 (0.66) \\
\hline
disgust &9, 10, 17 & 4 (0.31),  \hspace{0.08cm} 24 (0.26) \\
\hline
\end{tabular}
}
\label{relationship}
\end{table}


\noindent\paragraph{Distribution Matching $\mathcal{L}_{DM}$:}
Here, we propose the distribution matching loss for coupling the expression and AU tasks.
The objective is to align the predictions of the expression and AU tasks by ensuring consistency between them.
From expression predictions we create AU pseudo-predictions and  match these with the network's actual AU predictions. For instance, if the network predicts \emph{happy} with probability 1, but also predicts that AUs 4, 15 and 1 are activated (which are associated with \emph{sad} according to Table \ref{relationship}), this discrepancy is corrected through the loss function, which infuses prior knowledge into the network to guide consistent predictions.

For each sample $x$, the expression predictions $p_{expr}$ are represented as the softmax scores over the seven basic expressions, while the AU activations $p_{AU}$ are represented as the sigmoid scores over $17$ AUs. 
We  then match the distribution over AU predictions $p_{AU_i}$ with a distribution 
$q_{AU_i}$, where the AUs are modeled as a mixture over the basic expression categories: 
\begin{equation}
    q_{AU_i} = \sum_{{expr}} p_{expr} \cdot p_{{AU_i}|{expr}}, 
\label{eq:distr}
\end{equation} 
where $p_{{AU_i}|{expr}}$ is deterministically defined from Table~\ref{relationship}, being 1 for prototypical or observational AUs, and 0 otherwise.  For example, AU2 is prototypical for  \emph{surprise} and observational for \emph{fear}, hence $q_{\text{AU2}} = p_{\text{surprise}} + p_{\text{fear}}$. 
So with this matching if, e.g., the network predicts \emph{happy} with probability 1 (i.e., $p_{\text{happy}}=1$), then only the prototypical and observational AUs of \emph{happy} (i.e., AUs 12, 25 and 6) need to be activated in the distribution q: $q_{\text{AU12}} = 1$; $q_{\text{AU25}} = 1$; $q_{\text{AU6}} = 1$, whereas the rest $q_{{AU}_i}$ are 0. 

The distributions $p_{AU_i}$ and $q_{AU_i}$ are then matched
by minimizing the binary cross-entropy loss term: 
\begin{align}
\mathcal{L}_{DM} = \mathbb{E} \Bigg[ \sum_{AU_i}[ -q_{AU_i} \cdot \text{log }p_{AU_i}] \Bigg] , \label{eq:coupleloss}
\end{align}
where all available train samples are used to match the predictions. \\

\noindent\textbf{Soft  co-annotation $\mathcal{L}_{SCA}$:} 
We also introduce a soft co-annotation loss to further couple the expression and AU tasks.
This loss generates soft expression labels that are guided by AU labels, infusing prior knowledge of their relationship. The soft labels are then matched with the expression predictions, which is particularly beneficial in cases of limited data with partial or no annotation overlap.


Given an image $x$ with ground truth AU annotations $y_{au}$, we first co-annotate it with a \emph{soft label} in the form of a  distribution over expressions and then match this label with the expression predictions $p_{expr}$.
For each basic expression, an indicator score $I_{expr}$ is computed based on the presence of its prototypical and observational AUs: 

\begin{align}
    I_{expr} =  \sum_{AU_i} w_{AU_i} \cdot y_{AU_i}  \biggl/ \sum_{AU_i} w_{AU_i}  
\end{align} 

\noindent Here, $w_{AU_i}$ is 1 if $AU_i$ is prototypical for $y_{expr}$ (from Table \ref{relationship}), $w$ if observational and 0 otherwise.
For example: $ I_{\text{happy}}  = (y_{\text{AU12}} + y_{\text{AU25}} + 0.51 \cdot y_{\text{AU6}}) / (1+1+0.51)$.
This indicator score is converted into a probability score over expression categories to form the \emph{soft} expression label $q_{expr}$:

\begin{align}
    q_{expr} &=  e^{I_{expr}} \biggl/ \sum_{{expr}'} e^{I_{{expr}'}}
\end{align}

Every image with ground truth AU annotations is assigned a \emph{soft} expression label, and the predictions $p_{expr}$ and are matched with these soft labels by minimizing the cross-entropy loss term:

\begin{align}
\mathcal{L}_{SCA} =  \mathbb{E} \Bigg[ \sum_{expr}[ -q_{expr} \cdot \text{log }p_{expr}] \Bigg] \label{eq:coupleloss2}
\end{align}

\begin{figure}[t]
\centering
\includegraphics[height=7cm]{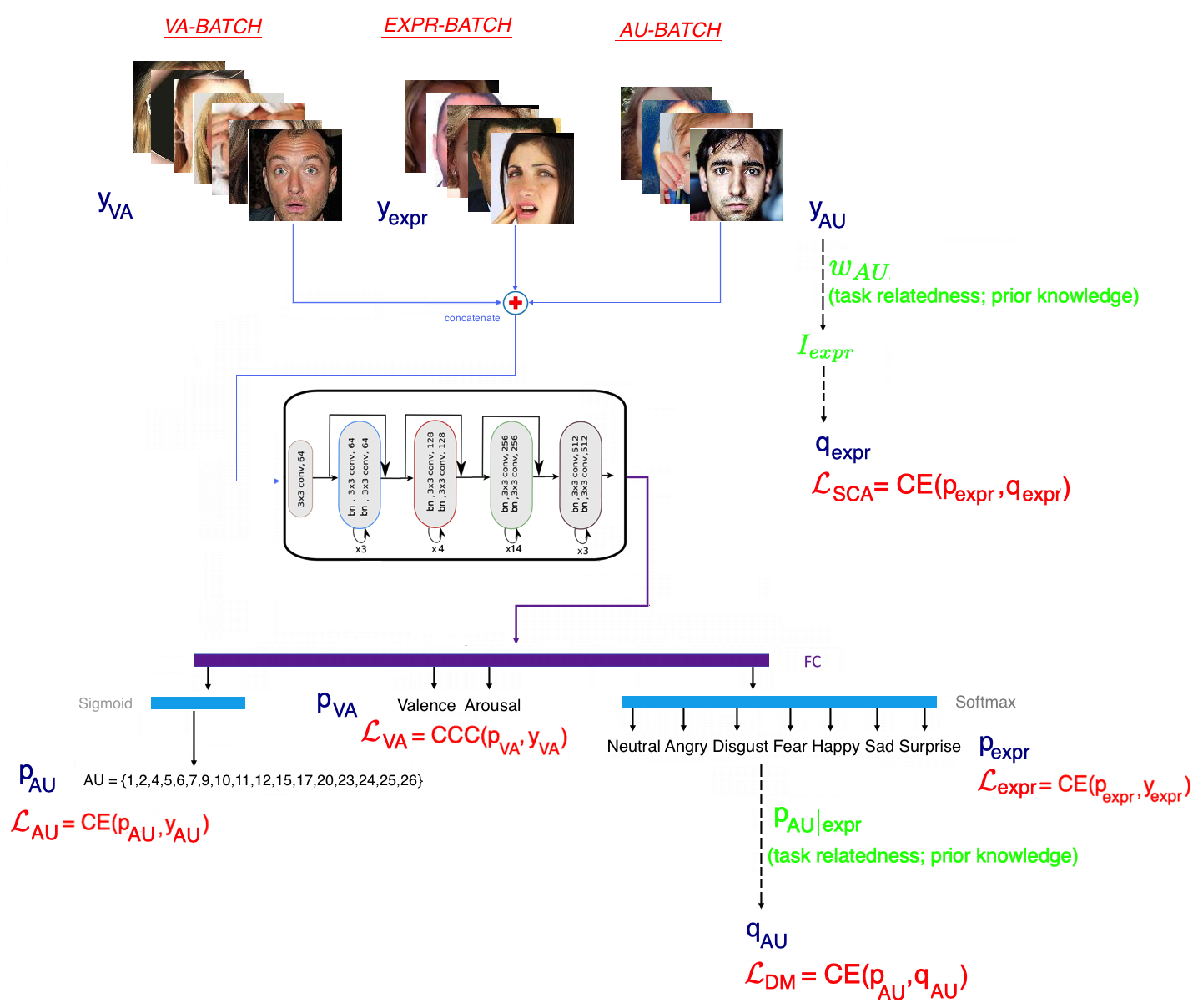} 
\caption{FaceBehaviorNet's architecture is designed to minimize an objective function during training, which includes loss terms associated with the three behavior tasks, as well as the proposed distribution matching and soft co-annotation losses.}
\label{facebehaviournet}
\end{figure}

The architecture of FacebehaviourNet, illustrated in Fig. \ref{facebehaviournet}, is structured around residual units, with 'bn' indicating batch normalization layers;   convolution layer being in the format: filter height $\times$ filter width 'conv.', number of output feature maps; the stride being equal to 2 only on convolutional layers with filters $1 \times 1$. The model integrates the VA estimation, 7 basic expression recognition, and 17 AU detection tasks within the same embedding space derived from a shared feed-forward layer.

\paragraph{Databases}
The \textbf{Aff-Wild} database \cite{kollias2019deep,zafeiriou2017aff} contains around 300 videos with 1.25M frames annotated in terms of VA.
The \textbf{AffectNet} database \cite{mollahosseini2017affectnet} contains around 400K images manually annotated for 7 basic expressions (plus contempt) and VA. 
The \textbf{RAF-DB} database \cite{li2017reliable} contains around 15K facial images annotated for 7 basic expressions. 
The \textbf{EmotioNet} database \cite{fabian2016emotionet} contains  950K automatically annotated images and 50K manually annotated images for 11 AUs. 
The \textbf{DISFA} database \cite{mavadati2013disfa} consists of 27 videos, each of which has around 5000 frames, where each frame is coded with the AU intensity on a six-point discrete scale. There are in total 12 AUs. 
The \textbf{BP4D-Spontaneous} database\cite{zhang14bp4d} (denoted as BP4D) contains 61 subjects with around 225K frames annotated for the occurrence and intensity of 27 AUs. This database has been used as a part of the FERA 2015 Challenge \cite{valstar2015fera}, in which only 11 AUs (1,2,4,6,7,10,12,14,15,17,23) were used. 
The \textbf{BP4D+} database \cite{zhang2016multimodal} is an extension of BP4D incorporating different modalities as well as more subjects. It is annotated for occurrence of 34 AUs and intensity for 5 of them. It has been used as a part of the FERA 2017 Challenge \cite{valstar2017fera}, in which only 10 AUs were used. 



Recent studies \cite{hu2024bridging, hu2024rethinking} have highlighted inconsistencies in database partitioning and evaluation practices in ABA, leading to unfair comparisons. To address this, a unified partitioning protocol was proposed, incorporating demographic information to ensure fairness and comparability. It was shown that methods previously considered sota may not perform as well under this new protocol. In this work, we adopt this updated partitioning protocol for the AffectNet, RAF-DB, EmotioNet, and DISFA datasets, which were re-annotated and repartitioned  (referred to as 'New').

\paragraph{Performance Measures}
For the overall performance, we use: i) the  CCC for evaluating VA estimation on Aff-Wild and AffectNet; CCC takes values in $[-1, 1]$;  ii) the F1 score for evaluating expression recognition on RAF-DB and AffectNet; iii) the F1 score for evaluating AU detection on DISFA, EmotioNet, BP4D and BP4D+; F1 score takes values in $[0, 1]$. Higher values are desirable for all metrics.

For calculating fairness (with respect to age, gender and race), we use: i) the fairness CCC (fCCC) on AffectNet (for Aff-Wild, no demographic labels exist); ii) the Equality of Opportunity (EOP) on RAF-DB and AffectNet; iii) the Equal Opportunity Difference (EOD) on EmotioNet and DISFA (for BP4D and BP4D+ no demographic labels exist). Lower values are desirable for all these fairness metrics; both EOP and EOD take values in $[0, 1]$, with values in $[0, 0.1]$ indicating fair methods.
More details about all evaluation metrics exist in the supplementary.

\section{Experimental Results}

 Training implementation details and ablation studies are included in the supplementary material.
 Experimental results and analysis for face detection  and localisation is included in the supplementary material.





\subsection{Overall Facial Behaviour Analysis}

\noindent
\textbf{BER} Table \ref{comparison_fer} provides a comprehensive performance comparison (in terms of F1 score) for 7 Basic Expression Recognition on AffectNet and RAF-DB. The comparison includes our proposed toolkit, the state-of-the-art (sota) methods, and other existing toolkits (py-feat and LibreFace). When utilizing the original database splits for model training and evaluation, our toolkit demonstrates a superior performance, exceeding the sota by at least 1.8\% on AffectNet and 2.4\% on RAF-DB. Moreover, our toolkit significantly outperforms the other toolkits, with improvements of at least 12.1\% on AffectNet and 10.5\% on RAF-DB.
When adopting the new database splits proposed by \cite{hu2024bridging, hu2024rethinking}, our toolkit continues to lead in performance, surpassing the sota by margins of at least 2.4\% on AffectNet and 2.5\% on RAF-DB. Notably, the performance gains over the sota are even more pronounced when using the new database splits compared to the original splits.

\begin{table}[ht]
\caption{Performance evaluation (in terms of F1 score in \%) of 7 basic expression recognition by FacebehaviourNet, the state-of-the-art and other toolkits} 
\label{comparison_fer}
\centering
\scalebox{.95}{
\begin{tabular}{ |c||c|c|c|c| }
 \hline
\multicolumn{1}{|c||}{\begin{tabular}{@{}c@{}} Databases  \end{tabular}} 
 & \multicolumn{2}{c|}{\begin{tabular}{@{}c@{}}  AffectNet \end{tabular}}  & \multicolumn{2}{c|}{RAF-DB}  \\
 \hline
  &\begin{tabular}{@{}c@{}} Original \end{tabular}  & New  & \begin{tabular}{@{}c@{}}  Original \end{tabular}  & New  \\ 

  \hline
 \hline
DAN   &   65.7 & 60.0  & 64.5 &  70.9 \\
\hline
EAC      & 65.3 & 60.3  & 63.6 & 75.5  \\
\hline
MA-Net & 64.5 & 55.4 & 60.7 & 65.2  \\
\hline
EfficientFace   & 63.7   & 58.6  &  68.7 &  73.2   \\
\hline
\hline

py-feat   & 55.0 & - & 49.3 & - \\
\hline
LibreFace & 49.7 & -  & 59.6 & - \\ 
\hline

\hline
\hline
\begin{tabular}{@{}c@{}} \textbf{FacebehaviourNet}  \end{tabular}    & \textbf{67.1} & \textbf{62.4} &  \textbf{71.1}  &  \textbf{78.0}  \\ 
\hline
\end{tabular}
}
\end{table}

\noindent
\textbf{VA-E} Table \ref{comparison_va} provides a comprehensive performance comparison (in terms of CCC score) for VA Estimation on Aff-Wild and AffectNet. The comparison includes our proposed toolkit and the sota methods (as we mentioned in the Related Work section, no existing toolkit performs this task). When utilizing the original database splits for model training and evaluation, our toolkit demonstrates a superior performance, exceeding the sota by at least 4.9\% on Aff-Wild and 4.5\% on AffectNet. 
When utilizing the new split for AffectNet proposed by \cite{hu2024bridging, hu2024rethinking} \footnote{Aff-Wild, BP4D and BP4D+ do not contain demographic information and have not been restructured by \cite{hu2024bridging, hu2024rethinking}}, our toolkit maintains its leading performance, exceeding the sota by at least 4.1\%.

\begin{table}[ht]
\caption{Performance evaluation (in terms of CCC in \%) of VA Estimation by FacebehaviourNet and the state-of-the-art} 
\label{comparison_va}
\centering
\scalebox{.83}{
\begin{tabular}{ |c||c|c|c| }
 \hline
\multicolumn{1}{|c||}{\begin{tabular}{@{}c@{}} Databases  \end{tabular}} 
 & \multicolumn{1}{c|}{\begin{tabular}{@{}c@{}}  Aff-Wild \end{tabular}}  & \multicolumn{2}{c|}{AffectNet}  \\
 \hline
  &\begin{tabular}{@{}c@{}} Original \end{tabular}  & \begin{tabular}{@{}c@{}}  Original \end{tabular}  & New  \\ 

  \hline
 \hline
FUXI   &  52.0  & 56.5  & 74.0   \\
\hline
SITU   & 53.1   & 57.5   &  71.1   \\
\hline
CTC &  50.2  & 52.3  &71.0  \\
\hline
AffWildNet & 50.0   &  - & - \\
\hline
VA-StarGan &  50.2   & 54.5  & 72.4 \\
\hline
MT-EmotiEffNet & 51.7  & 57.2 &  71.8 \\

\hline
\hline
\begin{tabular}{@{}c@{}} \textbf{FacebehaviourNet}  \end{tabular}    & \textbf{58.0} &  \textbf{62.0}  & \textbf{78.1}  \\ 
\hline
\end{tabular}
}
\end{table}

\noindent
\textbf{AUD} Table \ref{comparison_au} provides a comprehensive performance comparison (in terms of F1 score) for AU Detection on EmotioNet, DISFA, BP4D and BP4D+. The comparison includes our proposed toolkit, the sota methods, and other existing toolkits (py-feat, LibreFace, and OpenFace). When utilizing the original database splits for model training and evaluation, our toolkit demonstrates a superior performance, exceeding the sota by at least 2.2\% on EmotioNet, 10.2\% on DISFA, 22\% on BP4D and 4.3\% on BP4D+. Moreover, our toolkit significantly outperforms the other toolkits, with improvements of over 19.3\% on EmotioNet, 1.3\% on DISFA, 23\% on BP4D and 6.4\% on BP4D+.
When adopting the new database splits proposed by \cite{hu2024bridging, hu2024rethinking} $^{\textcolor{red}{2}}$, 
our toolkit continues to exhibit leading performance, surpassing the sota by at least 1.7\% on EmotioNet and 1.3\% on DISFA. It is important to note that the top-performing sota method across all databases (with the exception of BP4D) is AUNets, an ensemble method comprising 90 models. Despite this, our toolkit consistently outperforms AUNets in every case examined.

\begin{table}[ht]
\caption{Performance evaluation (in terms of F1 in \%) of AU Detection by FacebehaviourNet, the state-of-the-art and other toolkits} 
\label{comparison_au}
\centering
\scalebox{.7}{
\begin{tabular}{ |c||c|c|c|c|c|c| }
 \hline
\multicolumn{1}{|c||}{\begin{tabular}{@{}c@{}} Databases  \end{tabular}} 
 & \multicolumn{2}{c|}{\begin{tabular}{@{}c@{}}  EmotioNet \end{tabular}}  & \multicolumn{2}{c|}{DISFA} & \multicolumn{1}{c|}{BP4D} & \multicolumn{1}{c|}{BP4D+}  \\
 \hline
  &\begin{tabular}{@{}c@{}} Original \end{tabular}  & New  & \begin{tabular}{@{}c@{}}  Original \end{tabular}  & New & Original & Original  \\ 

  \hline
 \hline
Res50   &   44.0 & 64.3  & 48.9 & 41.1 & 50.0 & 45.0  \\
\hline
FUXI   &  50.2  & 77.9  & 49.8 & 43.7 & 63.2 & 56.2  \\
\hline
SITU   &  49.7  & 77.2  & 49.2 & 40.4 & 62.7 & 55.8 \\
\hline
CTC   &  47.6  &  74.6 & 51.1 & 45.9 & 61.1 & 54.6 \\
\hline
ME-GraphAU  & 49.8 & 72.9  & 52.3 & 43.0  & 65.5 &  56.7  \\
\hline
AUNets & 53.6 & 82.8  & 54.1  & 51.8 &  63.0  & 57.7  \\
\hline

\hline
\hline

py-feat   & 36.5 & - & 54.0 & -  & 61.4  & 52.4 \\
\hline
LibreFace & 35.7 &  - & 63.0  & - & 62.0 & 55.6 \\ 
\hline
OpenFace & 35.5 & -  & 59.0 & -  & 61.3 & 53.5 \\ 
\hline

\hline
\hline
\begin{tabular}{@{}c@{}} \textbf{FacebehaviourNet}  \end{tabular}    &  \textbf{55.8} & \textbf{84.5} &  \textbf{64.3} & \textbf{53.1}  & \textbf{85.0}  & \textbf{62.0}  \\ 
\hline
\end{tabular}
}
\end{table}





\subsection{Fairness Behaviour Analysis}

\noindent
\textbf{BER} 
Table \ref{comparison_fer_fairness} presents a detailed fairness comparison across various demographic attributes (age, race, and gender) for BER on AffectNet and RAF-DB. This comparison involves the same methods as those outlined in Table \ref{comparison_fer}. The results indicate that our toolkit consistently demonstrates greater fairness across all demographic attributes and databases when compared to both sota and the other two toolkits.
Notably, on both databases utilized, the following observations can be made. For \emph{gender}, our toolkit is unbiased, with an EOP below 10\%, whereas the sota and other toolkits exhibit bias.  For \emph{race}, although our toolkit is not entirely fair, the EOP score of approximately 15\% suggests it is close to meeting fairness criteria, in contrast to the sota and other toolkits, which are biased with EOP scores of 20\% or higher. For \emph{age}, our toolkit does not meet fairness criteria, as the EOP scores exceed 26\%.

\begin{table}[ht]
\caption{Fairness evaluation for demographic attributes (age, gender and race) (in terms of EOP in \%) of 7 basic expression recognition by FacebehaviourNet, the state-of-the-art and other toolkits} 
\label{comparison_fer_fairness}
\centering
\scalebox{.8}{
\begin{tabular}{ |c||c|c|c|c|c|c| }
 \hline
\multicolumn{1}{|c||}{\begin{tabular}{@{}c@{}} Databases  \end{tabular}} 
 & \multicolumn{3}{c|}{\begin{tabular}{@{}c@{}}  AffectNet \end{tabular}}  & \multicolumn{3}{c|}{RAF-DB}  \\
 \hline
  & Age & Gender & Race &  Age & Gender & Race \\ 

  \hline
 \hline
DAN   & 31.6 & 15.4 & 22.1 & 32.1 & 14.1  & 21.1\\
\hline
EAC      &  31.9 & 12.5 & 20.2 & 32.5  & 13.1  & 21.6 \\
\hline
MA-Net & 32.5 & 13.5 & 18.9 & 33.6 & 14.5  & 22.3 \\
\hline
EfficientFace   & 32.1 & 13.1 & 19.5 & 33.4 & 14.0  & 21.5 \\
\hline
\hline

py-feat  & 33.4 & 13.3 & 21.1 & 28.3 & 13.8 & 20.9   \\
\hline
LibreFace & 35.4 & 13.4 & 22.6 & 30.1 & 14.2 & 22.4   \\
\hline
\hline
\begin{tabular}{@{}c@{}} \textbf{FacebehaviourNet}  \end{tabular}    & \textbf{28.0}  & \textbf{9.6}  & \textbf{15.1} & \textbf{26.5} &  \textbf{9.0}  &  \textbf{15.1}  \\ 
\hline
\end{tabular}
}
\end{table}

\noindent
\textbf{VA-E} 
Table \ref{comparison_va_fairness} presents a detailed fairness comparison on AffectNet across various demographic attributes (age, race, and gender) for VA-E. This comparison involves the same methods as those outlined in Table \ref{comparison_va}. The results indicate that our toolkit consistently demonstrates greater fairness across all demographic attributes compared to  sota.

\begin{table}[ht]
\caption{Fairness evaluation for each demographic attribute (age, gender and race) (in terms of fCCC in \%) of VA Estimation by FacebehaviourNet and the state-of-the-art. $\downarrow$ score is better.} 
\label{comparison_va_fairness}
\centering
\scalebox{.85}{
\begin{tabular}{ |c||c|c|c| }
 \hline
AffectNet & \multicolumn{1}{|c||}{\begin{tabular}{@{}c@{}} Age  \end{tabular}} 
 & \multicolumn{1}{c|}{\begin{tabular}{@{}c@{}}  Gender \end{tabular}}  & \multicolumn{1}{c|}{Race}  \\

  \hline
 \hline
FUXI   &  56.7  & 44.0  &  52.1  \\
\hline
SITU   &  53.1  &  43.4  &   48.3  \\
\hline
CTC &  60.3  & 47.2  &  55.6 \\
\hline
VA-StarGan &  57.8   & 46.4  & 53.1 \\
\hline
MT-EmotiEffNet & 54.8  &  43.7 & 49.2  \\

\hline
\hline
\begin{tabular}{@{}c@{}} \textbf{FacebehaviourNet}  \end{tabular}    & \textbf{50.2} &  \textbf{39.5}  & \textbf{45.1}  \\ 
\hline
\end{tabular}
}
\end{table}

\noindent
\textbf{AUD} 
Table \ref{comparison_au_fairness} presents a detailed fairness comparison across various demographic attributes (age, race, and gender) for AUD on EmotioNet and DISFA. This comparison involves the same methods as those outlined in Table \ref{comparison_au}. Consistent with the results observed for BER, our toolkit demonstrates superior fairness across all demographic attributes and databases compared to the sota and the 3 toolkits. The findings for \emph{race} and \emph{age} align with those observed in the BER analysis.
The primary difference arises in the case of \emph{gender}, where our toolkit remains unbiased (EOP below 10\%) across both databases, while the sota and other toolkits are unbiased on EmotioNet but exhibit bias on DISFA.

\begin{table}[ht]
\caption{Fairness evaluation for each demographic attribute (age,
gender and race) (in terms of EOD in \%) of AU detection  by FacebehaviourNet, the state-of-the-art methods and other toolkits} 
\label{comparison_au_fairness}
\centering
\scalebox{.7}{
\begin{tabular}{ |c||c|c|c|c|c|c| }
 \hline
\multicolumn{1}{|c||}{\begin{tabular}{@{}c@{}} Databases  \end{tabular}} 
 & \multicolumn{3}{c|}{\begin{tabular}{@{}c@{}}  EmotioNet \end{tabular}}  & \multicolumn{3}{c|}{DISFA}   \\
 \hline
& Age & Gender & Race & Age & Gender & Race \\ 

  \hline
 \hline
Res50    &  41.6  & 6.5 & 23.1  & 50.4 & 15.6 & 39.6 \\
\hline
FUXI    &  35.8  & 8.0 & 22.2  & 48.3 & 19.5 & 49.3 \\
\hline
SITU    &  40.8  & 7.8 & 22.0 & 45.7 & 12.8 &  33.3\\
\hline
CTC    &  41.6  & 6.8 & 21.8  & 42.8 & 12.5  & 32.3\\
\hline
ME-GraphAU  &  37.1  & 6.4 & 20.0 & 41.2 & 17.8 & 45.1 \\
\hline
AUNets  &  39.9  & 6.9 & 19.8 & 48.3 & 19.6 & 40.2 \\
\hline

\hline
\hline

py-feat    & 38.7   & 8.9  & 17.3 & 40.3 & 13.5  & 29.6 \\
\hline
LibreFace  &  42.4  & 11.07 & 19.3 & 43.6 & 15.6  & 31.6 \\
\hline
OpenFace  &  45.5  & 8.0 &  20.6 & 52.1 & 13.9 & 32.7 \\
\hline

\hline
\hline
\begin{tabular}{@{}c@{}} \textbf{FacebehaviourNet}  \end{tabular}    &  \textbf{33.9} & \textbf{5.3} & \textbf{15.1} & \textbf{33.5} & \textbf{9.1} & \textbf{19.8} \\
\hline
\end{tabular}
}
\end{table}


\subsection{Generalizability and Downstream Tasks}

\noindent
\paragraph{Generalisability} 
The exceptional generalization performance of our toolkit across the test sets of the seven databases used in its training serves as a strong indicator of its effectiveness and versatility. To further demonstrate and validate the robustness and quality of the features learned by our toolkit, we show that it can generalize its knowledge and capabilities to unseen affect recognition databases that were not utilized during its training and that possess different statistical properties and contexts.
Table \ref{comparison_cross_db} provides a comprehensive performance comparison for BER on Aff-Wild2 (in terms of F1) \cite{kollias2024sam2clip2sam,zafeiriou1,Tailor,springer,kollias20247th,gerogiannis2024covid,kollias2024domain,kollias2023multi,kollias2024distribution,psaroudakis2022mixaugment,kollias2021distribution,kollias2019face,kollias2019deep,kollias2019expression,kollias2022abaw,kollias2023abaww,kollias2023btdnet,kollias2023facernet,kollias2023multi,kollias20246th,kollias2024distribution,kollias2024domain,kolliasijcv,zafeiriou2017aff,hu2024bridging,psaroudakis2022mixaugment,kollias2020analysing,kollias2021distribution,kollias2021affect,kollias2019face,kollias2021analysing,kollias2023ai,kollias2023deep2,arsenos2023data,gerogiannis2024covid,salpea2022medical,arsenos2024uncertainty,karampinis2024ensuring,arsenos4674579nefeli,miah2024can,arsenos2024commonn,cis}, for AUD on GFT (in terms of F1) and for VA-E on AFEW-VA (in terms of CCC). The comparison includes our proposed toolkit, the sota methods (AffWildNet, JAA-NET and FUXI) -that have been trained on one of these databases-, and an existing toolkit (AFFDEX 2.0). Our toolkit outperforms FUXI by 5\%, AFFDEX 2.0 by 3.5\%, AffWildNet by 15\% and JAA-Net by 7\%.

\begin{table}[ht]
\caption{Performance evaluation (in \%) between FacebehaviourNet and  state-of-the-art on 3 databases not utilized in its training}
\label{comparison_cross_db}
\centering
\scalebox{.78}{
\begin{tabular}{ |c||c|c|c| }
 \hline
\multicolumn{1}{|c||}{\begin{tabular}{@{}c@{}} Databases \end{tabular}} & Aff-Wild2-Expr &  \multicolumn{1}{c|}{AFEW-VA} & \multicolumn{1}{c|}{GFT}  \\
 \hline
 & F1  &\multicolumn{1}{c|}{\begin{tabular}{@{}c@{}} CCC \end{tabular}}  &\multicolumn{1}{c|}{\begin{tabular}{@{}c@{}} F1  \end{tabular}}   \\
  \hline
  \hline
JAA-Net \cite{jaanet}  & - & - & 55.0 \\
\hline
AffWildNet & -  & 54.0  & - \\
\hline
FUXI & 34.5  & - & -\\
\hline
\hline
AFFDEX 2.0 \cite{bishay2023affdex} & 36.0  & - & -\\
\hline
\hline
\begin{tabular}{@{}c@{}} \textbf{FacebehaviourNet}  \end{tabular}   &  \begin{tabular}{@{}c@{}} \textbf{39.5} \end{tabular} & \textbf{69.0}  &   
\textbf{62.0}  \\
\hline
\end{tabular}
}
\end{table}

\paragraph{Downstream Task: Compound Expression Recognition}
Here, we perform new experiments and utilize our toolkit as a foundation model, because it has learned good features that encapsulate all aspects of facial behaviour. By leveraging this foundation model, we aim to capitalize on its ability to generalize across various tasks and domains, thus enabling more efficient and effective transfer learning. 
Specifically, we conduct zero-shot and few-shot learning experiments on the downstream task of Compound Expression Recognition (CER) to evaluate the model's adaptability and performance with minimal or no additional task-specific training data. These experiments not only demonstrate the model's robustness and versatility in handling new, unseen tasks, but also highlight the potential for reducing the need for large labeled datasets in some specialized applications.

For the zero-shot learning experiment, we use the predictions of our toolkit together with the rules  from \cite{du2014compound} to  generate  compound  expression (CE)  predictions.
We compute a candidate score,  $\mathcal{C}_{expr}$, for each compound expression: 

\begin{align}
\mathcal{C}_{expr} &= \mathcal{I}_{AU} + \mathcal{F}_{expr} + \mathcal{D}_{VA} \\
 \mathcal{I}_{AU} &=  [\sum_{AU_i} p_{{AU_i}|{expr}}] ^ {-1} \cdot  \sum_{AU_i} p_{AU_i} \cdot p_{{AU_i}|{expr}} 
\nonumber\\
\mathcal{F}_{expr} &= p_{expr_1} + p_{expr_2} 
\nonumber\\
\mathcal{D}_{VA}  &=\left\{  
\begin{array}{ll} 
      1, &  p_{V} > 0\\
      0, & \text{otherwise} \\
\end{array} 
\right. \nonumber
\end{align}

\noindent
$\mathcal{I}_{AU}$ expresses our toolkit's AU predictions that are associated with this CE according to \cite{du2014compound}. 
%
$\mathcal{F}_{expr}$  expresses our toolkit's predictions of the BE $expr_1$ and $expr_2$ that form the CE (e.g., if  \emph{happily surprised}, then $expr_1$ is \emph{happy} and $expr_2$ is \emph{surprise}).
$\mathcal{D}_{VA}$ is added only to the \emph{happily surprised} and \emph{happily disgusted} expressions and is either 0 or 1 depending on whether our toolkit's valence prediction is negative or positive, respectively. 
The final prediction is the class that obtained the maximum candidate score.

Table \ref{comparison_zero_few_shot} provides a comprehensive performance comparison for CER on EmotioNet (in terms of F1) and RAF-DB (in terms of AA). Our toolkit consistently outperforms all sota by significant margins in both zero- and few-shot settings. As anticipated, the best overall performance is achieved by our toolkit under the few-shot setting.

\begin{table}[ht]
\caption{Performance evaluation (in \%) on CER between FacebehaviourNet and the state-of-the-art; 'AA' is the average accuracy
}
\label{comparison_zero_few_shot}
\centering
\scalebox{0.75}{
\begin{tabular}{ |c||c|c| }
 \hline
\multicolumn{1}{|c||}{\begin{tabular}{@{}c@{}} Databases \end{tabular}} & \multicolumn{1}{c|}{EmotioNet} & \multicolumn{1}{c|}{RAF-DB}  \\
 \hline
  & F1  &AA  \\
  \hline
  \hline
NTechLab \cite{emotionet2016}  & 25.5 & - \\
\hline
VGG-FACE-mSVM \cite{li2017reliable} & -  & 31.6 \\
\hline
DLP-CNN \cite{li2017reliable} & - & 44.6 \\
\hline
\hline
\begin{tabular}{@{}c@{}} \textbf{zero-shot  FacebehaviourNet}  \end{tabular} & \textbf{31.2}    &  \textbf{46.7}   \\
\hline
\begin{tabular}{@{}c@{}} \textbf{fine-tuned FacebehaviourNet} \end{tabular} &  \textbf{39.3}  & \textbf{55.3}    \\
\hline
\end{tabular}
}
\end{table}


\subsection{Efficiency Analysis}

We compare the computation cost of our toolkit (for predicting the 3 behaviour tasks) to the other toolkits (LibreFace, OpenFace and py-feat).  
Further details on the settings exist in the supplementary.
Table \ref{comparison_flops} presents a comparison in terms of  FPS, total number of parameters (in millions) and total number of  GFLOPs.
From Table \ref{comparison_flops}, we observe that our toolkit is of almost similar efficiency to that of LibreFace (almost same total number of parameters and GFLOPs). Our toolkit provides more accurate VA, basic expression and AU analysis whilst running at least 1.9 times faster than OpenFace and py-feat. It is important to highlight that our reported efficiency metrics reflect the performance of our toolkit when simultaneously predicting VA, AUs, and basic expressions, whereas LibreFace and py-feat predict only AUs and basic expressions concurrently, and OpenFace is limited to predicting AUs only.

\begin{table}[ht]
\caption{Efficiency (in terms of FPS) and model size comparison (in terms of FLOPs, total number of parameters and size) between FacebehaviourNet and other toolkits}
\label{comparison_flops}
\centering
\scalebox{.8}{
\begin{tabular}{ |c||c|c|c|c| }
 \hline
Toolkits & \# Params (M) & GFLOPs & FPS  \\
 \hline
  \hline
LibreFace    &  22.5 & 3.7  & 26.9 \\
OpenFace  & 44.8 & 7.4 &   13.5 \\
py-feat   & 49.3 & 8.2  & 12.2 \\
\hline
\hline
\begin{tabular}{@{}c@{}} \textbf{FacebehaviourNet}  \end{tabular}   & \textit{23.1} & \textit{3.8} & \textit{26.2} \\
\hline
\end{tabular}
}
\end{table}


\section{Conclusion}
In this paper, we introduced Behavior4All, an open-source toolkit for in-the-wild Face Localization, Valence-Arousal Estimation, Basic Expression Recognition, and Action Unit Detection within a single framework. Behavior4All is shown to surpass all state-of-the-art methods and the existing toolkits in overall performance, fairness and generalizability, while also being computationally efficient.

{\small
\bibliographystyle{ieee_fullname}
\bibliography{egbib}
}

\end{document}


\title{Behaviour4All: in-the-wild Facial Behaviour Analysis Toolkit}

\author{First Author\\
Institution1\\
Institution1 address\\
{\tt\small firstauthor@i1.org}
\and
Second Author\\
Institution2\\
First line of institution2 address\\
{\tt\small secondauthor@i2.org}
}
\maketitle


\section{Databases Utilized}

Here, we provide more details on the databases that have been utilized for the development (training and testing) of Behaviour4All toolkit.

The \textbf{Aff-Wild} database \cite{kollias2018deep,zafeiriou2017aff} has been the first large scale captured in-the-wild database, containing 298 videos (200 subjects) of around 1.25M frames, annotated in terms of valence-arousal that range in $[-1,1]$. It served as benchmark for the Aff-Wild Challenge organized in CVPR 2017. The training set consists of around 1M frames and the test set of around 216K. 
Given that Aff-Wild is a video database where consecutive frames often exhibit identical or very similar VA values, and considering that FacebehaviourNet is a CNN model that does not leverage temporal dependencies between frames, we adopted a frame-sampling strategy during training. Specifically, for each retained frame in the training videos, we skipped the subsequent four frames to reduce redundancy and focus on more distinct samples.

The \textbf{AFEW-VA} dataset \cite{kossaifi2017afew} consists of 600 video clips (without audio) totalling to around 30K frames that were annotated in terms of valence-arousal; the values are discrete in the range of [-10,10]. The values have been scaled to [-1,1] so as to be consistent with the other datasets.

The \textbf{AffectNet} database \cite{mollahosseini2017affectnet} contains around 400K images manually annotated for 7 basic expressions (plus contempt) and VA that ranges in $[-1,1]$.
The training set of this database consists of around 321K images and the validation of 5K. The validation set is balanced across the different expression categories. 
In this work, we exclude the contempt class and perform data cleaning by removing images for which there was a mismatch between the VA and expression labels. In more detail: i) images annotated as neutral should have radius of the valence-arousal vector smaller than 0.15; ii)  images annotated as sad or disgusted or fearful should have negative valence; iii) images annotated as angry should have negative valence and positive arousal; iv) images annotated as happy should have positive valence

The \textbf{RAF-DB} database \cite{li2017reliable} contains 12.2K training and 3K test facial images annotated in terms of the 7 basic expressions. It further contains a small subset of around 3k images annotated for 11 compound expressions. 

The \textbf{EmotioNet} database \cite{fabian2016emotionet} contains around 1M images and was released for the EmotioNet Challenge in 2017. 950K images were automatically annotated and the remaining 50K images were manually annotated with 11 AUs (1,2,4,5,6,9,12,17,20,25,26); around half of the latter constituted the validation and the other half the test set of the Challenge. Additionally, a subset of about 2.5K images was annotated with the 6 basic and 10 compound emotions.

The \textbf{DISFA} database \cite{mavadati2013disfa} is a lab controlled database consisting of 27 videos each of which has 4,845 frames, where each frame is coded with the AU intensity on a six-point discrete scale. AU intensities equal or greater than 2 are considered as occurrence, while others are treated as non-occurrence. There are in total 12 AUs (1,2,4,5,6,9,12,15,17,20,25,26). 

The \textbf{BP4D-Spontaneous} database\cite{zhang14bp4d} (in the rest of the paper we refer to it as BP4D) contains 61 subjects with 223K frames and is annotated for the occurrence and intensity of 27 AUs. There are 21 subjects with 75.6K images in the training, 20 subjects with 71.2K images in the  development and 20 subjects with 75.7K images in the test partition. This database has been used as a part of the FERA 2015 Challenge \cite{valstar2015fera}, in which only 11 AUs (1,2,4,6,7,10,12,14,15,17,23) were used. 

The \textbf{GFT} database consists of 96 videos of 96 subjects totalling around 130K frames. It is annotated for the occurrence of 10 AUs (1,2,4,6,10,12,14,15,23,24). The training set consists of 78 subjects of around 108K frames and the test set of 18 subjects of around 24.5K frames.

The \textbf{BP4D+} database \cite{zhang2016multimodal} is an extension of BP4D incorporating different modalities as well as more subjects (140). It is annotated for occurrence of 34 AUs and intensity for 5 of them. It has been used as a part of the FERA 2017 Challenge \cite{valstar2017fera}, in which only 10 AUs (1,4,6,7,10,12,14,15,17,23) were used. There are 2952 videos of 41 subjects with 9 different views in the training set, 1431 of 20 subjects  with 9 different views in the validation set and 1080 videos of 30 subjects in the test set. 

The \textbf{Aff-Wild2} database \cite{kollias2019deep,kollias2019expression,kollias2022abaw,kollias2023abaww,kollias2023multi,kollias20246th,kollias2024distribution,kolliasijcv,zafeiriou2017aff,kollias2020analysing,kollias2021distribution,kollias2021affect,kollias2019face,kollias2021analysing} is the largest in-the-wild database and the only one to be annotated in a per-frame basis for the seven basic expressions, twelve action units (AUs 1,2,4,6,7,10,12,15,23,24,25, 26) and valence and arousal (in the range $[-1, 1]$). In total, it consists of 564 videos of around 2.8M frames with 554 subjects. In this work, we utilize the set of images annotated with the 7 basic expressions, which we denote as Aff-Wild2-Expr.

The \textbf{WIDER FACE} database \cite{yang2016wider} comprises around 33K images sourced from the internet from a variety of real-world scenarios, featuring a total of 395K annotated faces. The annotations include bounding boxes for each face. The WiderFace dataset is diverse, covering a broad range of face variations, including different scales, poses, occlusions, expressions, and lighting conditions. The dataset is split into training
(40\%), validation (10\%) and testing (50\%) subsets by randomly sampling from 61 scene categories. The dataset is  divided into three subsets —Easy, Medium, and Hard—based on the difficulty level of detecting the faces. The Easy subset includes faces that are relatively large, well-lit, and unobstructed, while the Medium subset presents more challenging conditions, such as smaller faces or mild occlusions. The Hard subset is particularly demanding, with faces that are extremely small, heavily occluded, or captured under poor lighting conditions.

The \textbf{UFDD} database \cite{nada2018pushing} comprises 6.5K images, with a total of 11K annotated faces. The annotations are precise, with bounding boxes provided for each face. The images in UFDD are selected to represent a wide variety of challenging scenarios, including severe occlusions, extreme poses, exaggerated facial expressions, varying image quality, and challenging lighting conditions (including both overexposure and underexposure). Moreover, UFDD includes images with faces captured in different environments and contexts, such as crowded scenes, group photos, and images with background clutter, which are often missing in other datasets. 


\section{Performance Metrics}

Here we provide an overview and more detailed explanation on the various metrics that we utilize to measure both the overall performance and the fairness of behaviour analysis performed by FaceBehaviourNet (component of our Behaviour4All toolkit), the various sota methods and the other existing toolkits (OpenFace, LibreFace, py-feat and AFFDEX 2.0). We also present and explain the performance metrics used to measure the face detection and landmark localization performance of FaceLocalizationNet, the sota methods and the other toolkits.\\

\noindent
\textbf{Basic Expression Recognition} 
\\

\noindent
1) \underline{$\text{F}_{1}$ Score}: It evaluates the overall expression recognition performance across the entire partition (training, validation, or test) by averaging the $\text{F}_{1}$ scores of each expression class $c \in C$. It ranges from $[0, 1]$, with higher values being more desirable. It is defined as:
\begin{equation}
\smash{
\text{F}_{1} = \frac{1}{|C|} \sum_{c} F_1^c}
\end{equation}
where:  
$\text{F}_1^c$ is the $\text{F}_{1}$ Score for the $c$-th expression class across the entire partition. \\

\noindent
2) \underline{Equality of Opportunity (EOP)}: 
It evaluates the fairness of basic expression recognition with respect to a demographic attribute (age, race, gender). It ensures that the model performs equally well across different subgroups of a demographic attribute. This prevents the model from favouring one subgroup over another, thus promoting fairness. It ensures that one can quantify and reduce disparities in the model's performance across different subgroups.
It takes values in $[0,1]$, with 0 indicating perfect equality of opportunity; values equal or below 0.1 are considered fair.

For each subgroup $g$ of a demographic attribute $G$, at first we compute the confusion matrix $C_g$. We then normalize each of these matrices so that the sum of each row is 1, turning them into error rate matrices. This normalization gives the error rate of predicting class $j$ when the  true class is $i$. A normalised confusion matrix is defined as:
\begin{equation}
\hat{C}_g [i,j] = \frac{C_g [i,j]}{\sum_j C_g [i,j]}  \nonumber 
\end{equation}

Then, we compute the pairwise distances between the normalized confusion matrices using the Mean Absolute Deviation (MAD). For matrices $\hat{C}_{g}$ and $\hat{C}_{g'}$ (of subgroups $g$ and $g'$): 
\begin{equation}
    \text{MAD} (\hat{C}_{g}, \hat{C}_{g'}) = \frac{ \sum_{i} \sum_{j} \left | \hat{C}_{g} [i,j] - \hat{C}_{g'} [i,j] \right |}{N^2} \nonumber
\end{equation}

Finally, we aggregate the pairwise distances to get a single measure of equality of opportunity across all subgroups. We use the mean of these distances. Therefore, EOP is defined as:
\begin{equation}
\text{EOP} = \frac{2}{|G| \cdot (|G| -1)} \sum_{g \in G} \sum_{g' \in G, g \ne g'} \text{MAD} (\hat{C}_{g}, \hat{C}_{g'}).
\end{equation}

\noindent
\textbf{AU Detection} 
\\

\noindent
1) \underline{$\text{F}_{1}$ Score}: It evaluates the overall performance of AU Detection across the entire partition (i.e., training, validation, or test set). It is calculated by averaging the $\text{F}_{1}$ scores of each AU over the whole partition. It takes values in $[0, 1]$, with higher values being desirable. \\

\noindent
2) \underline{Equal Opportunity Difference (EOD)}: 
It evaluates the fairness of AU detection with respect to a demographic attribute (age, race, gender). It measures fairness by ensuring that the True Positive Rates (TPR) for each AU are similar across different demographic subgroups. This helps to avoid biases where some subgroups might have significantly higher or lower TPRs for certain AUs, leading to unfair advantages or disadvantages.
It takes values in $[0, 1]$; values equal or below 0.1 are considered fair.

For each subgroup $g \in G$ and each $au$, at first, we compute the: 
\begin{equation}
    \text{TPR}_{g}^{\text{au}} = \frac{\sum_{i \in G_k} \mathds{1} \left ( \hat{y}_i^{au} = 1 \land y_i^{au} = 1 \right )}{\sum_{i \in G_k} \mathds{1} \left ( y_i^{au} = 1 \right ) }  \nonumber
\end{equation}

\noindent
where:  $G_k$ is the set of indices of samples belonging to subgroup $g$; $\hat{y}_i^{au}$ is the prediction for the specific $au$ for sample $i$; $y_i^{au}$ is the true label for the specific $au$ for sample $i$. 

Then, for each $au$, we calculate the difference between the maximum and minimum TPRs across all subgroups; finally, EOD is the average of these differences across all $M$ AUs:
\begin{equation}
\text{EOD} = \frac{1}{M} \sum_{au} \left ( \max_{g \in G} \text{TPR}_g^{au} - \min_{g \in G} \text{TPR}_g^{au} \right ) .
\end{equation} \\

\noindent
\textbf{Valence-Arousal Estimation} 
\\

\noindent
1) \underline{CCC}: It evaluates the overall performance of VA Estimation across the entire partition (i.e., training, validation, or test
set). It takes values in $[-1, 1]$, with higher values being desirable. It is calculated as:
\begin{equation}
\text{CCC} = \frac{1}{2} \left ( \text{CCC}_\text{V} + \text{CCC}_\text{A}  \right )
\end{equation}
where:\\
$ \text{CCC}_i = 2 \cdot s_{xy} / \left[ s_x^2 + s_y^2 + (\bar{x} - \bar{y})^2 \right]$, \\
with $i =$ V or A; $s_x$ and $s_y$ represent the variances of the valence/arousal annotations and predicted values, respectively; $\bar{x}$ and $\bar{y}$ denote the mean values of the corresponding annotations and predictions; $s_{xy}$ represents their covariance. \\

\noindent
2) \underline{fairness CCC (fCCC)}: It evaluates the fairness of VA Estimation with respect to a demographic attribute (age, race, gender). It evaluates the alignment between predicted values and annotations across diverse demographic subgroups within a demographic attribute. It is computed by averaging the mean CCC values across all subgroups:
\begin{equation}
\smash{
\text{fCCC} = \frac{1}{2 \cdot |G|} \sum_g  \left ( \text{CCC}_{\text{V}}^g + \text{CCC}_{\text{A}}^g  \right )}
\end{equation}
where: $\text{CCC}_i^g$ denotes the CCC
of valence/arousal calculated over all images of the $g$-th subgroup.\\

\noindent
\textbf{Face Detection} 
\\
The evaluation of face detection models on the WiderFace dataset is performed using the  Average Precision and mean Average Precision (mAP) metric, which is derived from the Precision-Recall (PR) curve. For each of the three difficulty settings (Easy, Medium, and Hard), at first, Precision (P) and Recall (R) are calculated at various confidence thresholds, then the PR curve is plotted by varying the detection confidence threshold, plotting Precision against Recall; finally, the Average Precision is calculated as the area under the PR curve (AUC). This is done for each difficulty setting, resulting in three AP values corresponding to the Easy, Medium, and Hard settings; their average creates the mAP.

\section{Training Implementation Details}

To address the presence of approximately 20\% tiny faces in the WIDER FACE training set, we adopt the approach outlined in previous works and randomly crop square patches from the original images and resize them to a resolution of $640 \times 640$ pixels during training. In addition to random cropping, we augment the training dataset by applying random horizontal flipping and photometric color distortion. FaceLocalizationNet is trained using SGD optimizer with a momentum of 0.9, a weight decay of 0.0005, and a batch size of 32 across four NVIDIA Tesla P40 GPUs, each with 24GB of memory. The learning rate is initially set to 
$10^{-3}$ and is increased to $10^{-2}$. The training process is concluded after 80 epochs.

The training of FaceBehaviorNet was performed in an end-to-end manner, using the SGD optimizer with 0.9 momentum and a learning rate of $10^{-4}$. The input images to FaceBehaviorNet are resized to dimensions $112 \times 112 \times 3$ and the pixel intensity values to $[-1, 1]$.
Training was performed on a Tesla V100 32GB GPU; training time was about 2 days.

\section{Experimental Results}

\subsection{Face Detection}
Table \ref{comparison_fd} provides a comprehensive performance comparison (in terms of AP) for Face Detection on WIDER FACE. The comparison includes: i) the FaceLocalizationNet component of our proposed toolkit; we show 4 cases: when the backbone network is a MobileNet-0.25 (denoted as FaceLocalizationNet (M-0.25)); when the backbone network is a quantized MobileNet-0.25 (denoted as FaceLocalizationNet-Q (M-0.25)); when the backbone network is a ResNet50 (denoted as FaceLocalizationNet (R50)); when the backbone network is a quantized ResNet50 (denoted as FaceLocalizationNet-Q (R50)); ii) the sota methods: MTCNN \cite{zhang2016joint}, ScaleFace \cite{yang2017face}, CMS-RCNN \cite{zhu2017cms}, MSCNN \cite{cai2016unified}, HR \cite{hu2017finding}, SSH \cite{najibi2017ssh}, Face R-CNN \cite{wang2017face} and FAN \cite{wang2017face}; and iii) online RetinaFace implementations either in Tensorflow (T) or Pytorch (P) (peteryuX-P-R50, biubug6-P-R50, biubug6-P-M0.25).  
It is evident that FaceLocalizationNet-Q (R50) and FaceLocalizationNet (R50) outperform all other sota methods, as well as their online implementations. 
Finally, let us note that the results that we provide for our models are performed in single scale, whereas most sota methods employ flipping as well as multi-scale strategies and box voting.

\begin{table}[ht]
\caption{Evaluation of Face Detection (in terms of AP in \%) on WIDER FACE dataset  by FaceLocalizationNet  and the state-of-the-art methods and other implementations} 
\label{comparison_fd}
\centering
\scalebox{.9}{
\begin{tabular}{ |c||c|c|c| }
 \hline
\multicolumn{1}{|c||}{\begin{tabular}{@{}c@{}} Method  \end{tabular}} 
 & \multicolumn{3}{c|}{\begin{tabular}{@{}c@{}}  WIDER FACE \end{tabular}}   \\
 \hline
& Easy & Medium & Hard \\ 

  \hline
 \hline
MTCNN    &  84.8  & 82.5 & 59.8  \\
\hline
ScaleFace & 86.8 & 86.7 & 77.2\\
\hline
CMS-RCNN & 89.9 & 87.4 & 62.4 \\
\hline
MSCNN    &  91.6  & 90.3 & 80.2   \\
\hline
HR & 92.5 & 91.0 & 80.6 \\
\hline
SSH & 93.1 & 92.1 & 84.5 \\
\hline
Face R-CNN & 93.7 & 92.1 & 83.1 \\
\hline
SFD & 93.7 &  92.5 & 85.9 \\
\hline
Face R-FCN & 94.7 & 93.5 & 87.4 \\
\hline
FAN & 95.2 & 94.0 & 84.1 \\
\hline
\hline
peteryuX-P-R50 &	95.5 & 94.0 &	84.4\\
\hline
biubug6-P-R50 & 95.5 &	94.0&	84.4 \\
\hline
biubug6-P-M0.25 & 	88.7 &	87.1 &	81.0 \\
\hline
\hline
\begin{tabular}{@{}c@{}} \textbf{FaceLocalizationNet (M-0.25)}  \end{tabular}    &  \textbf{93.5} & \textbf{92.0} & \textbf{79.8} \\
\hline
\begin{tabular}{@{}c@{}} \textbf{FaceLocalizationNet-Q (M-0.25)}  \end{tabular}     &  \textbf{93.5} & \textbf{91.9} & \textbf{79.6} \\
\hline
\begin{tabular}{@{}c@{}} \textbf{FaceLocalizationNet (R50)}  \end{tabular}    &  \textbf{95.6} & \textbf{94.6} & \textbf{88.5} \\
\hline
\begin{tabular}{@{}c@{}} \textbf{FaceLocalizationNet-Q (R50)}  \end{tabular}    &  \textbf{95.6} & \textbf{94.5} & \textbf{88.3} \\
\hline

\end{tabular}
}
\end{table}

\paragraph{Generalizability}

Table \ref{comparison_fd_gen} provides a comprehensive performance comparison for Face Detection on UFDD (in terms of mAP). The comparison includes the FaceLocalizationNet component of our proposed toolkit (the four cases that we described previously) trained on the WIDER FACE and the sota methods. Although all four versions of our model have not been trained on UFDD, they were able to surpass all sota methods that have been trained on  UFDD.

\begin{table}[ht]
\caption{Evaluation of Face Detection (in terms of mAP in \%) on UFDD dataset  by FaceLocalizationNet  and the state-of-the-art methods} 
\label{comparison_fd_gen}
\centering
\scalebox{1.}{
\begin{tabular}{ |c||c| }
 \hline
\multicolumn{1}{|c||}{\begin{tabular}{@{}c@{}} Method  \end{tabular}} 
 & \multicolumn{1}{c|}{\begin{tabular}{@{}c@{}}  UFDD \end{tabular}}   \\
 \hline
& mAP \\ 

  \hline
 \hline
Faster-RCNN & 52.1  \\
\hline
SSH & 695 \\
\hline
S3FD    &  72.5  \\
\hline
HR-ER & 74.2 \\
\hline
\hline
\begin{tabular}{@{}c@{}} \textbf{FaceLocalizationNet (M-0.25)}  \end{tabular}    &  \textbf{76.9} \\
\hline
\begin{tabular}{@{}c@{}} \textbf{FaceLocalizationNet-Q (M-0.25)}  \end{tabular}     &  \textbf{76.4} \\
\hline
\begin{tabular}{@{}c@{}} \textbf{FaceLocalizationNet (R50)}  \end{tabular}    &  \textbf{92.1}  \\
\hline
\begin{tabular}{@{}c@{}} \textbf{FaceLocalizationNet-Q (R50)}  \end{tabular}    &  \textbf{91.7}  \\
\hline

\end{tabular}
}
\end{table}

\subsection{Ablation study for FaceBehaviourNet}

Here, we compare the performance of FaceBehaviourNet when trained with the two extra losses, i.e.,  the Distribution Matching and Soft Co-Annotation losses of eq. 3 and 6 of the main manuscript vs: i) when trained in a normal multi-task learning manner (i.e., without using these two extra losses), ii) when trained using only the Soft Co-Annotation  loss, and iii) when trained using only the Distribution Matching loss. Table \ref{comparison_losses} shows the results for all these cases.

It can be observed that when FaceBehaviourNet is trained either with only the the Distribution Matching loss, or with only the  Soft Co-Annotation loss, or with these two extra losses. it displayed a better performance than when trained in a normal multi-task learning manner (i.e., without using these two extra losses). This holds on all databases across all metrics. 
Additionally, across all databases, best results have been achieved when FaceBehaviourNet was trained with both Soft Co-Annotation and Distribution Matching losses.

\begin{table*}[h]
\caption{Performance evaluation of valence-arousal, seven basic expression and action units predictions on all used databases provided by the FaceBehaviorNet when trained with/without coupling, under two task relatedness scenarios; 'AFA' is the average between the F1 Score and the Accuracy; 'AA' is the average accuracy}
\label{comparison_losses}
\centering
\scalebox{0.9}{
\begin{tabular}{ |c||c||c|c|c|c|c|c|c|c|c| }
 \hline
\multicolumn{1}{|c||}{\begin{tabular}{@{}c@{}} Databases \end{tabular}}   & \multicolumn{2}{c|}{Aff-Wild} & \multicolumn{3}{c|}{\begin{tabular}{@{}c@{}}  AffectNet \end{tabular}}   & \multicolumn{1}{c|}{RAF-DB} & \multicolumn{1}{c|}{EmotioNet} & \multicolumn{1}{c|}{DISFA}  & \multicolumn{1}{c|}{BP4D} & \multicolumn{1}{c|}{BP4D+}  \\
 \hline
 FaceBehaviorNet   
 &\multicolumn{1}{c|}{CCC-V}  
 &\multicolumn{1}{c|}{CCC-A} 
 &\multicolumn{1}{c|}{CCC-V} 
 &\multicolumn{1}{c|}{CCC-A} &\multicolumn{1}{c|}{\begin{tabular}{@{}c@{}}  F1  \end{tabular}}  
 & \multicolumn{1}{c|}{\begin{tabular}{@{}c@{}} F1 \end{tabular}} &\multicolumn{1}{c|}{\begin{tabular}{@{}c@{}}  F1  \end{tabular}} 
 & \multicolumn{1}{c|}{\begin{tabular}{@{}c@{}}  F1  \end{tabular}}
 &\multicolumn{1}{c|}{\begin{tabular}{@{}c@{}}  F1  \end{tabular}} 
 &\multicolumn{1}{c|}{\begin{tabular}{@{}c@{}}  F1  \end{tabular}} \\ 
  \hline
  \hline
  \begin{tabular}{@{}c@{}} typical MTL \\ (no extra loss) \end{tabular}
   & 52.0 & 38.0 & 56.0 & 50.0 & 59.0  & 67.0 & 46.0 & 54.0 & 76.0 & 56.0  \\
 \hline
 \hline
with Soft Co-Annotation   & 64.0 & 46.0 & 64.0  & 55.0 & 64.0    & 69.0  & 52.0  & 60.0  & 82.0 & 60.0  \\
 \hline
with  Distribution Matching   & 64.0  & 44.0 & 64.0  & 57.0 & 63.0    & 70.0  &53.0   & 58.0 &84.0  & 58.0  \\ 
 \hline 
\begin{tabular}{@{}c@{}}  \textbf{ with Soft Co-Annotation} \\ \textbf{and Distribution Matching}  \end{tabular}  & \textbf{68.0}   &  \textbf{48.0}   & \textbf{66.0} & \textbf{58.0}  & \textbf{67.1}      & \textbf{71.1}  &  \textbf{55.8}   & \textbf{64.3}  & \textbf{85.0}  & \textbf{62.0}   \\
 \hline 
\end{tabular}
}
\end{table*}

\subsection{Efficiency Analysis}

We ran all experiments with Intel i9-13900K CPU, 64GB DDR5 4800MHz RAM, and Nvidia GTX1080 on Windows 11 Operating System. E-cores and hyper-threading were disabled. We randomly picked 5000 images in a resolution of $224 \times 224$ and fed them to the existing toolkits and our toolkit and computed the average FPS, the total number of parameters (in millions) and the total number of Giga FLOPs. The results can be seen on Table 11 of the main manuscript.

{\small
\bibliographystyle{ieee_fullname}
\bibliography{egbib}
}